\documentclass{article}
\usepackage[a4paper]{geometry} 
\usepackage{graphicx} 
\usepackage{url}

\usepackage[style=apa]{biblatex}
\usepackage{hyperref}
\usepackage{booktabs, float}

\begin{document}

    \title{Can AI Help with Your Personal Finances?}

    \author{
        Oudom Hean\textsuperscript{*}, Utsha Saha, Binita Saha\\
        \textit{\{oudom.hean, utsha.saha, binita.saha\}@ndsu.edu}\\
        North Dakota State University, Fargo, ND, USA\\
        \textsuperscript{*}\textit{Corresponding author: Oudom Hean, oudom.hean@ndsu.edu}
        \\\\
        \small{Draft article, accepted for publication in \textit{\textbf{Applied Economics}} (Taylor \& Francis)}\\
        \small{Submitted on: July 29, 2024 } \\
        \small{Accepted on: December 30, 2024 }\\
        \small{DOI: \url{https://doi.org/10.1080/00036846.2025.2450384}}
    }

    \date{}

    \maketitle

    \section*{Abstract}
    In recent years, Large Language Models (LLMs) have emerged as a transformative development in artificial intelligence (AI), drawing significant attention from industry and academia. Trained on vast datasets, these sophisticated AI systems exhibit impressive natural language processing and content generation capabilities.
    This paper explores the potential of LLMs to address key challenges in personal finance, focusing on the United States. We evaluate several leading LLMs, including OpenAI’s ChatGPT, Google’s Gemini, Anthropic’s Claude, and Meta’s Llama, to assess their effectiveness in providing accurate financial advice on topics such as mortgages, taxes, loans, and investments. Our findings show that while these models achieve an average accuracy rate of approximately 70\%, they also display notable limitations in certain areas. Specifically, LLMs struggle to provide
    accurate responses for complex financial queries, with performance varying significantly across different topics. Despite these limitations, the analysis reveals notable improvements in newer versions of these models, highlighting their growing utility for individuals and financial advisors. As these AI systems continue to evolve, their potential for advancing AI-driven applications in personal finance becomes increasingly promising.
    \\

    \textbf{Keywords:} Large Language Models; Artificial Intelligence; Generative Pretrained Transformers; Personal Finances; Financial Literacy\\

    \textbf{JEL codes:} D14; G11; G53

    \section{Introduction}
    Large Language Models (LLMs), a significant development in artificial intelligence (AI), have emerged as a transformative technology, attracting substantial interest from both industry and academia. These advanced AI systems, trained on extensive datasets, have exhibited impressive capabilities across diverse domains, such as natural language
    processing, question answering, and content generation \parencite{Brown2020, Devlin2018}. As LLMs continue to evolve and advance, their potential applications, particularly in personal finance, are becoming increasingly relevant and practical, extending well beyond academic research.

    In this paper, we thoroughly investigate the potential of AI, specifically LLMs, in addressing a spectrum of personal finance issues \footnote{In this study, we use the terms AI and LLMs interchangeably. Technically, LLMs are a subset of AI, focusing on processing and
    generating human language.}. We systematically evaluate the responses of LLMs to a range of personal finance topics—such as mortgages, taxes, loans, and investments—within the United States context. We examine several prominent LLMs, including OpenAI's ChatGPT, Google's Gemini, Anthropic's Claude, and Meta/Facebook's Llama, to discern their comparative efficacy in delivering accurate and consistent financial advice.

    Our findings indicate that AI models, on average, correctly answer approximately 70 percent of posed questions. Notably, AI performance has shown consistent improvement over time. Older versions of these models answered only about 50-60 percent of questions correctly, while the latest models achieve accuracy rates up to 80 percent. ChatGPT and Claude emerged as the top performers among the various models tested, with Llama being the least accurate. Specifically, the latest versions of ChatGPT, such as ChatGPT 4o, and Claude 3.5 Sonnet, achieved accuracy rates exceeding 74 percent, whereas Llama3 70b maintained a lower accuracy rate of about 65 percent.

    Furthermore, our results demonstrate that these LLMs provide consistent answers when prompted with the same questions multiple times, alleviating concerns about inconsistent responses. These advancements in AI model performance have significant implications for future applications, particularly in the fields of natural language processing and question-answering systems within the finance industry and education
    sector.

    Personal finance is a critical aspect of everyone's life, yet it is often overlooked due to its complexity and the lack of accessible, personalized guidance \parencite{Lusardi2014}. Many people struggle with budgeting, investing, taxes, and financial planning, leading to suboptimal financial decisions and outcomes \parencite{Jappelli2013}. Traditionally, individuals have turned to financial advisors for help, but this can be expensive and only within reach for a few. Financial advisors usually charge fees, with robo-advisors at the more affordable end of this range and traditional in-person advisors at the higher end \parencite{Fisch2019}. These fees can mount up swiftly, making professional financial guidance a costly and out-of-reach service for many.

    Our study reveals that while AI currently faces limitations in assisting with personal finance, particularly with complex questions and variability in performance across topics, its potential for future applications is promising as technology advances. As AI models become more sophisticated, they could play an essential role in helping individuals manage their finances more effectively. One could use AI to provide tailored advice on various personal finance topics, such as budgeting, investments, loans, and tax planning. Moreover, AI has the potential to become an invaluable tool for financial advisors by improving their capability to analyze complex economic data, generate insights, and offer personalized recommendations to clients. The continued advancement of AI in this domain could lead to greater financial literacy and improved financial decision-making for both individuals and professionals.

    Our paper contributes to the growing body of literature analyzing the application of AI in finance and economics. While most studies focus on a limited set of AI models, they offer valuable insights into specific capabilities and limitations. For instance, \textcite{Niszczota2023} examine GPT-3.5 and GPT-4 models, evaluating their financial literacy to determine their potential as financial advisors for the public. Using a standardized financial literacy test on core topics like compound interest, tax-advantaged assets, and risk diversification, they find that GPT-4 achieves a near-perfect score of 99\%, a marked improvement over GPT-3.5’s 65-66\%. This advancement in GPT-4 indicates significant progress in understanding fundamental financial concepts and its potential accuracy in financial advisement.

    Further extending the capabilities of LLMs, \textcite{Kim2024} demonstrate that GPT-4 can outperform human analysts in predicting company earnings from financial statements, particularly in challenging scenarios. This study underscores the model's capacity for analyzing complex financial information and positioning LLMs as competitive with, and in some cases superior to, human expertise in financial prediction tasks. Similarly, \textcite{Dowling2023} and \textcite{Korinek2023} explore the broader potential of ChatGPT in financial and economic research, emphasizing LLMs’ expanding utility in these domains.

    Despite these advancements, other studies caution against over-reliance on LLMs for financial advisory roles. \textcite{Lakkaraju2023} investigate the reliability and fairness of LLMs in financial advisement by comparing ChatGPT and Bard with SafeFinance, a rule-based chatbot. While LLMs provide plausible responses, they often make substantial errors in retrieving financial information and show inconsistent accuracy across user demographics. In contrast, the more limited SafeFinance offers safer, traceable advice, suggesting that current LLMs may fall short in consistently reliable advisement, especially in high-stakes financial contexts.

    Other research has focused on AI’s role in dynamic financial markets. \textcite{Yu2023} introduce FINMEM, an LLM-based autonomous trading agent designed to enhance decision-making under volatile conditions. FINMEM integrates a profiling module that adjusts risk preferences, a layered memory system that processes time-sensitive financial data, and a decision-making module that synthesizes insights for trading actions. Their findings show that FINMEM can adapt to fluctuating market environments, demonstrating resilience and superior trading performance in complex scenarios. Similarly, \textcite{Ding2023} develop a Local-Global model for predicting outcomes in the Chinese A-share market by combining LLM-processed financial news with stock-specific features. This integration allows for more accurate market predictions, highlighting the potential of combining LLMs with market data for enhanced financial forecasting.

    Our study also draws on broader research that links technology, education, and economic outcomes \parencite{Goldin2009, Hean2022, Hean2024}. These studies underscore the transformative role of technology in shaping economic behavior and decision-making, offering a comprehensive backdrop for our work within the broader narrative of AI's impact on economies.

    The rest of the paper is organized as follows: Section \ref{sec2} discusses the data sets used in this study, and Section 3 outlines the methodology. Section 4 presents the findings, followed by a sensitivity analysis in Section 5. Finally, Section 6 presents the conclusion, discussing ethical considerations and offering suggestions for future research.

    \section{Data} \label{sec2}
    In this study, we utilize two distinct datasets: MoneyCounts and the National Financial Educators Council’s (NFEC) comprehensive suite of financial literacy tests. These datasets provide a broad range of topics and varying levels of expertise in personal finance understanding.

    MoneyCounts is a specialized financial literacy series developed by Penn State University’s Sokolov-Miller Family Financial and Life Skills Center \footnote{MoneyCounts’ data were retrieved on July 18, 2024, from \url{https://financialliteracy.psu.edu/explore-a-financial-topic}.}. It stands out for its extensive collection of practice quizzes designed to assess and improve financial literacy across diverse subjects. Key areas covered include the principles of financial literacy, financial planning after graduation, banking basics, budgeting essentials, car shopping tips, understanding credit cards, debt management strategies, environmental stewardship and finance, FICO credit scores, organizing financial clutter, financial knowledge specific to women, financial literacy for high school students, identity theft protection, insurance planning, money and nutrition, money and relationships, mortgage fundamentals, retirement planning, salary negotiation, saving and investing, smart financial goals, managing student loans, financial planning for study abroad, U.S. tax information for international and U.S. individuals, and understanding the time value of money.

    The NFEC dataset offers a detailed assessment of financial literacy through tests at beginner, intermediate, and advanced levels \footnote{NFEC’s data were retrieved on July 10, 2024 \url{https://www.financialeducatorscouncil.org/financial-literacy-test/}.}. These tests cover various topics, including financial psychology tests; savings, expenses, and budgeting tests; account management tests; loans and debt tests; credit profile tests; income tests; economic and government influences tests; risk management and insurance tests; investing and personal financial planning tests; and education and skill development tests.

    \section{Methodology}
    We assess the financial literacy performance of LLMs using the specified test sets and conduct an accuracy assessment to evaluate their effectiveness in delivering accurate and useful financial advice. Specifically, we employ zero-shot prompting, in which the LLMs are tested without prior exposure to specific examples. Our evaluation considers various aspects of personal finance, such as budgeting, investments, loans, and tax planning, to comprehensively analyze each model's capabilities. This evaluation includes a diverse range of LLMs from several leading providers. Specifically, we test the following models: OpenAI (i.e., ChatGPT 3.5, ChatGPT 4, and ChatGPT 4o); Google (i.e., Gemini and Gemini Advanced); Anthropic (i.e., Claude 3 Haiku, Claude 3.5 Sonnet, and Claude 3 Opus); and Meta/Facebook (i.e., Llama 3 8B and Llama 3 70B)\footnote{Llama models in this study are accessible through Groq - \url{https://groq.com/}}. Table \ref{tab:llm_models} shows the different AI models tested in the paper.

    \begin{table}[h!]
        \centering
        \caption{Large Language Models} \label{tab:llm_models}
        \resizebox{0.8\textwidth}{!}{
            \begin{tabular}{lccc}
                \toprule
                \textbf{Provider} & \textbf{Model}    & \textbf{Version}           & \textbf{License Type} \\ \midrule
                OpenAI            & ChatGPT 3.5       & gpt-3.5-turbo-1106         & Proprietary           \\
                OpenAI            & ChatGPT 4         & gpt-4-turbo                & Proprietary           \\
                OpenAI            & ChatGPT 4o        & gpt-4o                     & Proprietary           \\
                Google            & Gemini            & gemini-1.5-flash           & Proprietary           \\
                Google            & Gemini Advanced   & gemini-1.5-pro             & Proprietary           \\
                Anthropic         & Claude 3 Haiku    & claude-3-haiku-20240307    & Proprietary           \\
                Anthropic         & Claude 3.5 Sonnet & claude-3-5-sonnet-20240620 & Proprietary           \\
                Anthropic         & Claude 3 Opus     & claude-3-opus-20240229     & Proprietary           \\
                Meta/Facebook     & Llama 3 8B        & llama3-8b-8192             & Open Source           \\
                Meta/Facebook     & Llama 3 70B       & llama3-70b-8192            & Open Source           \\ \bottomrule
            \end{tabular}
        }
    \end{table}

    We initially assess the financial literacy of the models by asking each question once and evaluating the generated answer. However, it is well-known that generative AI can produce inconsistent answers even when asked the same questions multiple times. To address this, we conduct a sensitivity analysis by repeating these tests ten times. This approach allows us to evaluate the consistency and reliability of the AI models' responses to financial literacy questions more thoroughly.

    \section{Results}

    \subsection{Overall Performance of LLMs}

    Table \ref{tab:llms_performance} shows the overall performance of all AI models in addressing financial questions. On average, the models correctly answer approximately 68 percent of the questions, highlighting existing limitations in AI's personal financial literacy capabilities. However, the high standard deviation (i.e., 46 percent) suggests significant variability in performance; while some models perform well, others struggle with the same questions. These findings underscore that there is still room for improvement in AI’s role in personal finance.

    \begin{table}[h!]
        \centering
        \caption{Overall Performance of AI Models (Percentage of Correct Answers)} \label{tab:llms_performance}
        \resizebox{0.7\textwidth}{!}{
            \begin{tabular}{lccccc}
                \toprule
                \textbf{Model}    & \textbf{Observations} & \textbf{Mean} & \textbf{STD} & \textbf{Min} & \textbf{Max} \\ \midrule
                All models        & 5,540                 & 68.47         & 46.47        & 0            & 100          \\
                ChatGPT 3.5       & 554                   & 61.01         & 48.82        & 0            & 100          \\
                ChatGPT 4         & 554                   & 78.34         & 41.23        & 0            & 100          \\
                ChatGPT 4o        & 554                   & 74.91         & 43.39        & 0            & 100          \\
                Gemini            & 554                   & 64.8          & 47.8         & 0            & 100          \\
                Gemini Advanced   & 554                   & 71.3          & 45.28        & 0            & 100          \\
                Claude 3 Haiku    & 554                   & 61.73         & 48.65        & 0            & 100          \\
                Claude 3 Opus     & 554                   & 74.73         & 43.5         & 0            & 100          \\
                Claude 3.5 Sonnet & 554                   & 79.78         & 40.2         & 0            & 100          \\
                Llama3 8b         & 554                   & 53.43         & 49.93        & 0            & 100          \\
                Llama3 70b        & 554                   & 64.62         & 47.86        & 0            & 100          \\ \bottomrule
            \end{tabular}
        }
    \end{table}

    \textbf{Note:} The number of observations is the total number of questions in the test sets described in the text. The models received 100 percent for correct answers and zero for incorrect ones.

    While Table \ref{tab:llms_performance} highlights significant differences in response accuracy across various models, Figure 1 illustrates these results. Among these models, Claude 3.5 Sonnet stands out with the highest mean accuracy of 80 percent, indicating that it generally provides the most accurate responses. The latest versions of ChatGPT, specifically ChatGPT 4 and ChatGPT 4o, also perform well, with average correct answers of about 78 percent and 74 percent, respectively. However, these top-performing models exhibit relatively high standard deviations, suggesting some variability in their performance. On the other end of the spectrum, Llama3 8b demonstrates the lowest mean accuracy at 53 percent and the highest standard deviation at 50 percent, indicating that it struggles the most with providing accurate financial information.

    Finally, when comparing different versions of the same models, we generally observe a trend of improvement over time in LLMs. Each successive version demonstrates enhancements in accuracy and overall performance. In the sensitivity analysis, we will show that LLMs also improve in overall consistency over time. For example, newer iterations of models like ChatGPT and Claude have shown marked progress in providing correct responses more frequently and handling complex queries
    with greater reliability. This pattern suggests ongoing advancements in the underlying algorithms and training methodologies, contributing to more effective and efficient LLMs.

    \begin{figure}
        \centering
        \includegraphics[width=0.8\linewidth]{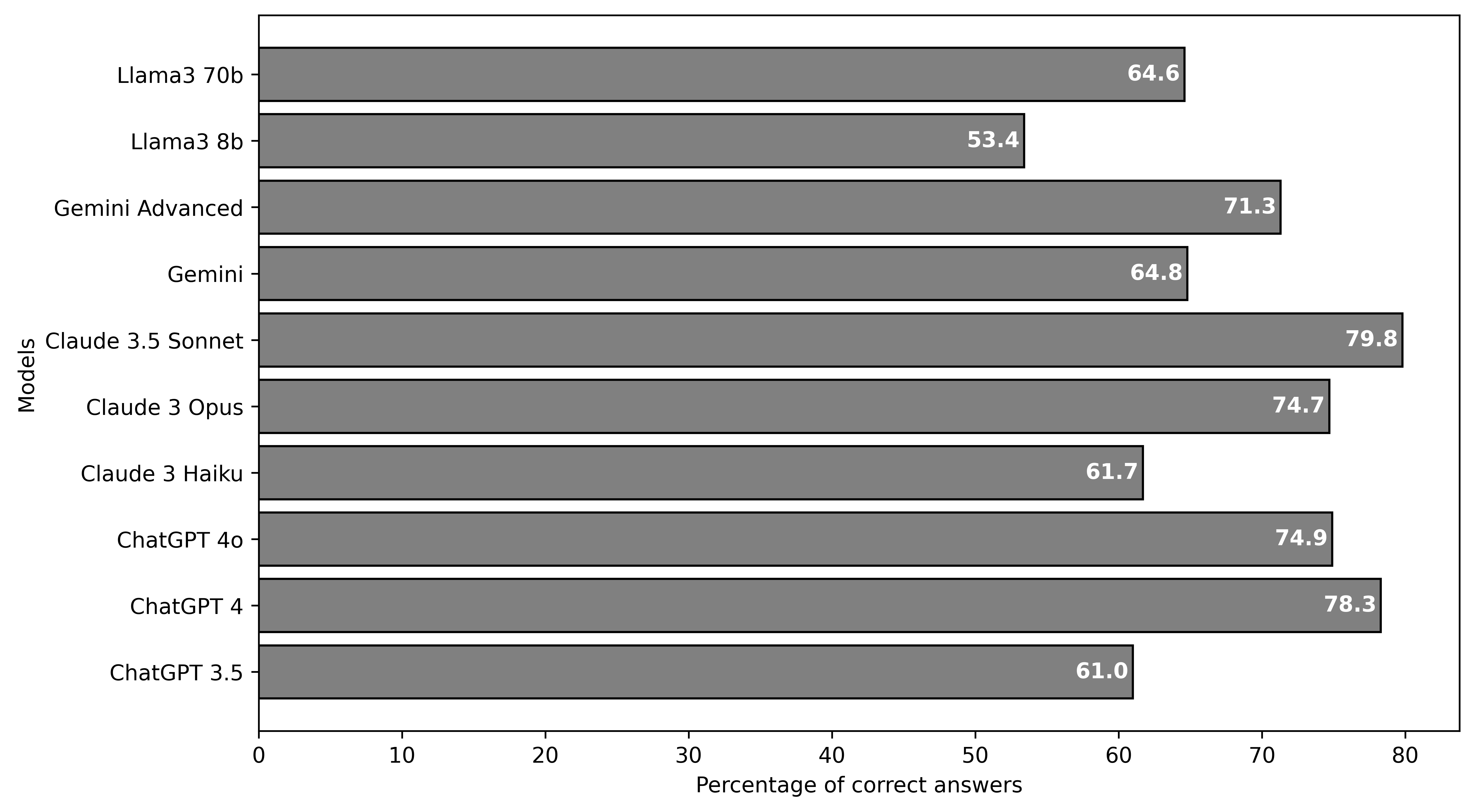}
        \caption{Performance of AI Models (Percentage of Correct Answers)}
        \label{fig:model_performance}
    \end{figure}

    \subsection{Performance of LLMs by Topic}

    Our data enables us to analyze the performance of different models across various topics. Table \ref{tab:ai_accuracy} presents the results of this analysis. We focus on a few topics that have attracted public attention. By examining these specific areas, we can highlight how different models perform and identify any notable trends or insights. This targeted discussion helps illustrate each model's strengths and weaknesses in real-world applications that matter to the public.

    First, understanding credit card concepts is crucial for individuals to make informed decisions about their financial well-being and to avoid potential pitfalls associated with credit card usage \parencite{Akinwande2024, Limbu2019}. The performance of large language models (LLMs) on credit card-related questions varies significantly. ChatGPT 4 is one of the top performers, correctly answering 80 percent of the questions. In contrast, Gemini, Claude 3 Haiku, and Llama 3 8B struggle, with only 40 percent of their answers being correct. The results suggest that while some models have a solid grasp of credit card concepts, others still face significant challenges.

    Second, women have made remarkable strides, becoming influential consumers and gaining social and professional positions. However, they still face numerous demands on their time and money, juggling home, work, and social priorities. Many women may face challenges in addressing financial goals throughout their lives and into retirement \parencite{Malhotra2010}. Therefore, it is crucial to provide tailored financial advice to women. Most AI models score in the 80-90 percent range, demonstrating their ability to provide relevant financial advice while navigating gender-specific considerations. This strong performance indicates that the models have been trained on a diverse dataset that includes information specific to women's financial needs. This is crucial given women's unique challenges in managing their finances and planning for the future.

    Third, money is a deeply personal and emotional topic, often more stressful to discuss than other aspects of life. In relationships, financial decisions can significantly impact various areas, from joint accounts to retirement planning \parencite{Olson2022}. AI models perform relatively well in addressing questions on ‘money and relationships,’ with most large language models achieving scores in the 70–100 percent range.

    Fourth, buying a house is arguably one of the most significant financial decisions, especially for Generation Z \parencite{Abdullah2024}. This decision involves numerous considerations, including affordability, location, and long-term financial commitments. LLMs perform moderately well when navigating the mortgage process, with accuracy rates ranging from 60 to 90 percent. AI could provide valuable guidance on mortgage options, interest rates, and repayment plans, helping prospective homeowners make informed decisions.

    Fifth, student loans, including federal and private options, are designed to help students cover post-secondary education costs \parencite{Gillen2015}. Navigating the complexities of borrowing and repayment strategies can be challenging for many students. AI models offer valuable assistance by helping students make informed decisions about borrowing strategies and repayment plans. Performance on student loan-related questions is generally good, with most models scoring in the 70-80 percent range. Therefore, LLMs could offer critical insights into interest rates, repayment options, and loan consolidation, aiding students in managing their financial responsibilities effectively and planning for a financially secure future.

    Sixth, taxes are a critical topic that everyone must understand to fulfill their legal and fiscal obligations, avoid penalties, and save money. Many individuals in the U.S. and internationally struggle with filing tax returns and paying taxes. Most AI models perform reasonably well on topics related to taxes for both domestic and international individuals, with accuracy rates ranging from 70 to 90 percent. These results indicate that AI has a fair capability to handle the intricacies of tax law, providing valuable support in managing tax-related responsibilities.

    Overall, the performance of LLMs across various financial topics showcases their potential as valuable tools for enhancing financial literacy and decision-making. Although some models still face challenges in some areas, the overall high accuracy rates indicate significant progress and reliability in handling complex financial topics. These findings highlight the importance of continued advancements in AI training methodologies to ensure diverse and inclusive datasets, ultimately making LLMs more effective in providing accurate, personalized financial guidance to a broad audience.

    \begin{table}[h!]
        \centering
        \caption{Accuracy of AI Models by Topic (Percentage of Correct Answers)}
        \label{tab:ai_accuracy}
        \resizebox{1\textwidth}{!}{
            \begin{tabular}{lcccccccccc}
                \toprule
                \textbf{Topic}                           & \textbf{ChatGPT} & \textbf{ChatGPT} & \textbf{ChatGPT} & \textbf{Gemini} & \textbf{Gemini} & \textbf{Claude 3.5} & \textbf{Claude 3} & \textbf{Claude 3} & \textbf{Llama3} & \textbf{Llama3} \\
                & \textbf{3.5}     & \textbf{4}       & \textbf{4o}      &                 & \textbf{Advanced} & \textbf{Haiku}      & \textbf{Opus} & \textbf{Sonnet} & \textbf{8B} & \textbf{70B} \\
                \midrule
                Finance Principles                       & 70               & 90               & 80               & 70              & 60                & 60                  & 70                & 70                & 60              & 70              \\
                Adult Learners                           & 50               & 100              & 80               & 60              & 90                & 60                  & 80                & 90                & 60              & 90              \\
                After Graduation                         & 60               & 90               & 70               & 60              & 80                & 60                  & 60                & 80                & 60              & 60              \\
                Banking                                  & 90               & 80               & 80               & 40              & 50                & 50                  & 60                & 60                & 40              & 50              \\
                Budgeting                                & 60               & 60               & 60               & 60              & 60                & 40                  & 30                & 50                & 50              & 50              \\
                Car Shopping                             & 50               & 80               & 80               & 80              & 80                & 70                  & 80                & 70                & 50              & 70              \\
                Credit Cards                             & 60               & 80               & 60               & 40              & 60                & 40                  & 50                & 70                & 40              & 50              \\
                Debt Management                          & 60               & 80               & 60               & 60              & 70                & 80                  & 80                & 80                & 60              & 70              \\
                Environmental Stewardship                & 100              & 100              & 90               & 90              & 100               & 100                 & 90                & 100               & 90              & 100             \\
                Fico Credit Score                        & 50               & 80               & 80               & 60              & 70                & 80                  & 90                & 90                & 60              & 50              \\
                Financial Clutter                        & 80               & 80               & 90               & 80              & 90                & 80                  & 90                & 90                & 80              & 90              \\
                Financial Literacy Overview              & 40               & 50               & 40               & 60              & 60                & 60                  & 60                & 60                & 40              & 50              \\
                For Women                                & 70               & 90               & 90               & 80              & 90                & 80                  & 90                & 90                & 50              & 80              \\
                High School Students                     & 60               & 70               & 70               & 60              & 60                & 60                  & 60                & 60                & 50              & 70              \\
                Identity Theft                           & 60               & 70               & 80               & 70              & 80                & 70                  & 70                & 90                & 70              & 70              \\
                Insurance Planning                       & 80               & 100              & 100              & 90              & 80                & 80                  & 90                & 90                & 80              & 90              \\
                Money \& Nutrition                       & 70               & 90               & 100              & 70              & 90                & 90                  & 80                & 90                & 80              & 70              \\
                Money \& Relationships                   & 80               & 80               & 90               & 60              & 70                & 60                  & 100               & 90                & 80              & 70              \\
                Mortgage                                 & 70               & 80               & 90               & 80              & 70                & 70                  & 90                & 90                & 70              & 60              \\
                Retirement Planning                      & 60               & 80               & 80               & 60              & 90                & 80                  & 80                & 80                & 60              & 80              \\
                Salary Negotiation                       & 70               & 70               & 70               & 60              & 80                & 50                  & 60                & 60                & 50              & 40              \\
                Saving \& Investing                      & 90               & 80               & 80               & 70              & 90                & 70                  & 70                & 90                & 70              & 70              \\
                SMART Goals                              & 50               & 60               & 60               & 60              & 50                & 50                  & 60                & 50                & 60              & 50              \\
                Student Loans                            & 70               & 80               & 80               & 70              & 90                & 80                  & 80                & 80                & 40              & 80              \\
                Study Abroad                             & 50               & 80               & 50               & 50              & 50                & 60                  & 60                & 70                & 50              & 60              \\
                Tax for International Individuals        & 80               & 80               & 80               & 60              & 50                & 70                  & 90                & 90                & 60              & 60              \\
                Tax for US Individuals                   & 70               & 70               & 80               & 60              & 80                & 70                  & 90                & 80                & 80              & 70              \\
                Time Value of Money                      & 60               & 80               & 70               & 70              & 70                & 70                  & 70                & 80                & 50              & 70              \\
                Account Management                       & 59.09            & 72.73            & 77.27            & 72.73           & 81.82             & 68.18               & 77.27             & 81.82             & 45.46           & 72.73           \\
                Credit Profile                           & 64.52            & 83.87            & 74.19            & 64.52           & 61.29             & 51.61               & 74.19             & 80.65             & 54.84           & 64.52           \\
                Economic \& Government Influences        & 48.15            & 81.48            & 81.48            & 66.67           & 81.48             & 70.37               & 88.89             & 92.59 & 33.33 & 70.37 \\
                Financial Psychology                     & 38.46            & 80.77            & 73.08            & 57.69           & 61.54             & 46.15               & 65.39             & 80.77             & 50              & 50              \\
                Higher Education                         & 70               & 85               & 80               & 80              & 80                & 70                  & 85                & 85                & 70              & 85              \\
                Income                                   & 60.87            & 73.91            & 60.87            & 60.87           & 69.57             & 39.13               & 65.22             & 78.26             & 47.82           & 52.17           \\
                Insurance                                & 66.67            & 85.18            & 74.07            & 81.48           & 74.07             & 66.67               & 77.78             & 85.18             & 48.15           & 70.37           \\
                Investing \& Personal Financial Planning & 41.31            & 71.74            & 69.57            & 50              & 67.39             & 45.66               & 78.26             & 78.26 & 30.43 & 52.17 \\
                Loans \& Debt                            & 60               & 70               & 70               & 50              & 60                & 50                  & 63.34             & 73.34             & 43.33           & 53.33           \\
                Savings, Expenses \& Budgeting           & 59.09            & 68.18            & 77.27            & 77.27           & 59.09             & 63.64               & 77.27             & 81.82             & 59.09 & 59.09 \\
                \bottomrule
            \end{tabular}
        }
    \end{table}

    \subsection{Performance of LLMs by Level of Complexity}
    The NFEC data categorizes each question into three difficulty levels: beginner, intermediate, and advanced. We use this dataset to evaluate the performance of various LLMs. By examining the results of this analysis, we can gain insights into the strengths and weaknesses of each model in handling questions of varying complexity. This detailed evaluation helps us understand which models are more effective in providing accurate responses based on the difficulty level of the questions.

    \begin{table}[h!]
        \centering
        \caption{Performance of AI by Level of Complexity (Percentage of Correct Answers)}
        \label{tab:ai_complexity}
        \resizebox{1\textwidth}{!}{
            \begin{tabular}{lcccccccccc}
                \toprule
                \textbf{Difficulty} & \textbf{ChatGPT} & \textbf{ChatGPT} & \textbf{ChatGPT} & \textbf{Gemini} & \textbf{Gemini} & \textbf{Claude 3} & \textbf{Claude 3} & \textbf{Claude 3.5} & \textbf{Llama 3} & \textbf{Llama 3} \\
                \textbf{Level}            & \textbf{3.5}     & \textbf{4}       & \textbf{4o}      &                 & \textbf{Advanced} & \textbf{Opus} & \textbf{Haiku} & \textbf{Sonnet} & \textbf{8B} & \textbf{70B} \\
                \midrule
                Beginner                  & 59.04            & 78.31            & 73.49            & 60.24           & 69.88             & 73.49             & 60.24             & 77.11               & 48.19            & 61.45            \\
                Intermediate              & 57.69            & 79.81            & 76.92            & 68.27           & 73.08             & 81.73             & 60.58             & 86.54               & 53.85            & 70.19            \\
                Advanced                  & 49.43            & 72.41            & 68.97            & 63.22           & 63.22             & 68.97             & 45.98             & 79.31               & 35.63            & 51.72            \\
                \bottomrule
            \end{tabular}
        }
    \end{table}

    \noindent\textit{Note: The models received 100 percent for correct answers and zero for incorrect ones.}

    Table~\ref{tab:ai_complexity} presents the findings of this analysis, highlighting how each model performs across the different difficulty levels. Overall, LLMs tend to struggle with questions at the advanced level. However, consistent with previous results, the latest versions of AI models show considerable improvement. Among the providers, the newest versions from OpenAI's ChatGPT and Anthropic's Claude achieve the best results, demonstrating their enhanced capability to handle more complex questions effectively.

    \section{Sensitivity Analysis}

    Generative AI may produce varying answers when the same questions are asked repeatedly. We perform a sensitivity analysis by conducting the tests ten times to mitigate this issue. This strategy allows us to assess the consistency and reliability of the AI models' responses to financial literacy questions more thoroughly.

    \begin{table}[h!]
        \centering
        \caption{Sensitivity Test (Percentage of Correct Answers)}
        \resizebox{0.7\textwidth}{!}{
            \label{tab:sensitivity_analysis}
            \begin{tabular}{lccccc}
                \toprule
                \textbf{Model}    & \textbf{Observations} & \textbf{Mean} & \textbf{STD} & \textbf{Min} & \textbf{Max} \\
                \midrule
                All models        & 55,400                & 68.12         & 44.08        & 0            & 100          \\
                ChatGPT 3.5       & 5,540                 & 60.02         & 45.11        & 0            & 100          \\
                ChatGPT 4         & 5,540                 & 77.85         & 39.29        & 0            & 100          \\
                ChatGPT 4o        & 5,540                 & 74.03         & 40.71        & 0            & 100          \\
                Gemini            & 5,540                 & 64.78         & 46.64        & 0            & 100          \\
                Gemini Advanced   & 5,540                 & 70.97         & 44.29        & 0            & 100          \\
                Claude 3 Haiku    & 5,540                 & 62.11         & 45.75        & 0            & 100          \\
                Claude 3 Opus     & 5,540                 & 74.62         & 41.24        & 0            & 100          \\
                Claude 3.5 Sonnet & 5,540                 & 79.57         & 38.97        & 0            & 100          \\
                Llama3 8B         & 5,540                 & 52.65         & 43.83        & 0            & 100          \\
                Llama3 70B        & 5,540                 & 64.57         & 46.69        & 0            & 100          \\
                \bottomrule
            \end{tabular}
        }
    \end{table}

    \noindent\textit{Note: In this analysis, each question in the data sets is repeated ten times.}

\begin{figure}[h!]
        \centering
        \includegraphics[width=0.8\textwidth]{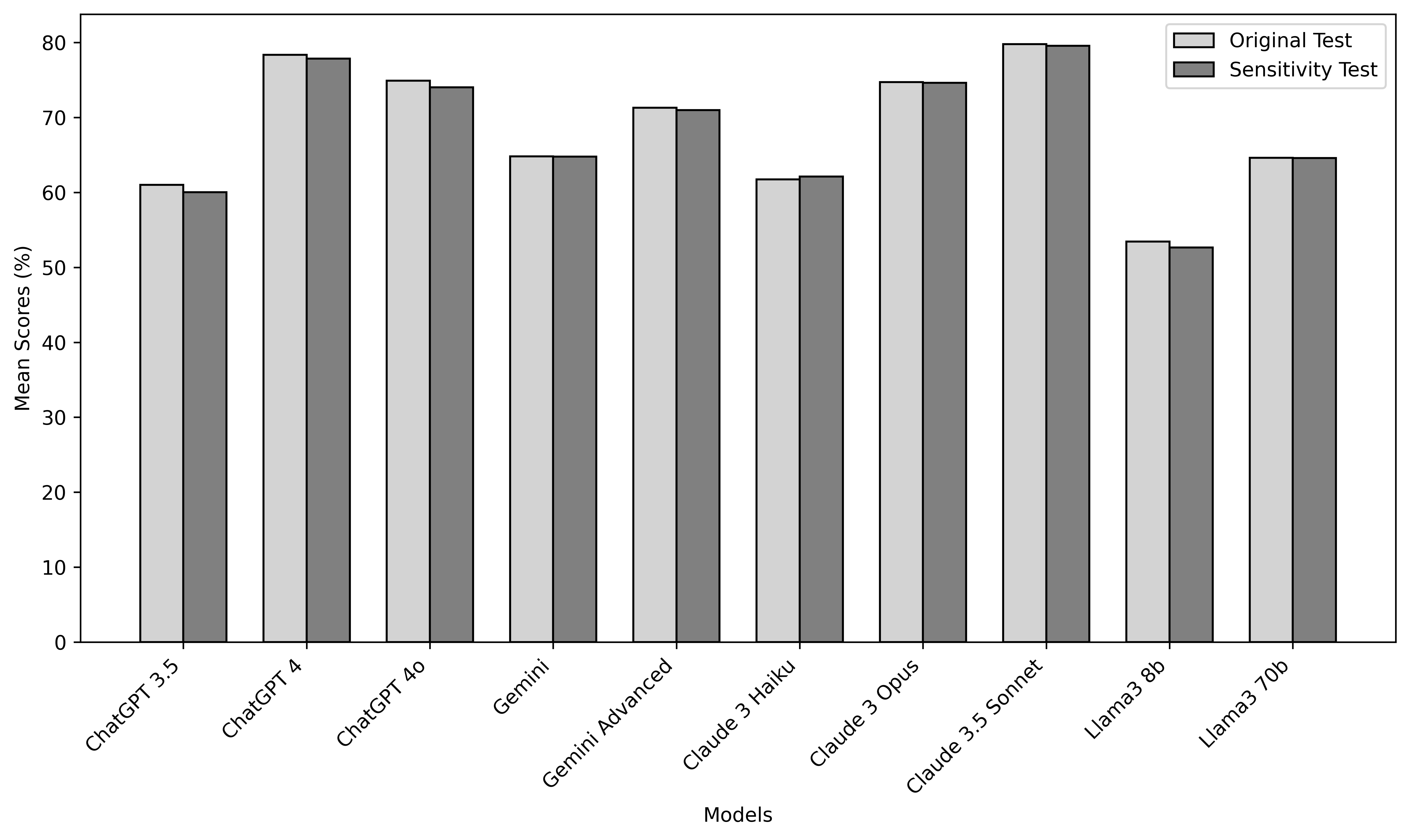}
        \caption{Original and Sensitivity Analysis Results}
        \label{fig:sensitivity_results}
    \end{figure}

    Table~\ref{tab:sensitivity_analysis} presents the findings from the sensitivity analysis, while Figure~\ref{fig:sensitivity_results} compares the original results and those obtained from the sensitivity analysis. Surprisingly, the results demonstrate a high level of robustness, indicating that the AI models' performance remains consistent and reliable despite repeated testing. This robustness underscores the reliability of the models in providing consistent responses to financial literacy questions.

    \noindent\textit{Note: For the sensitivity analysis, each question in the data sets is repeated ten times.}

    \section{Conclusions}

    In recent years, large language models have emerged as a transformative technology in artificial intelligence, garnering significant attention from both industry and academia. As LLMs evolve, their potential applications, particularly in personal finance, become increasingly relevant and practical, extending well beyond academic research.

    We investigate the potential of AI, specifically LLMs, in addressing a spectrum of personal finance issues. By examining a diverse array of personal finance topics—such as mortgages, taxes, loans, and investments—we systematically evaluate the responses provided by various AI models. We analyze several prominent LLMs, including OpenAI's ChatGPT, Google's Gemini, Anthropic's Claude, and Meta/Facebook's Llama, to assess their comparative efficacy in delivering accurate and actionable financial advice.

    Our findings indicate that AI models still have limitations in answering personal finance questions but have shown substantial improvement. Interestingly, all models provide strongly consistent results when the same questions are asked multiple times. As AI models become more sophisticated, they could play an essential role in helping individuals manage their finances more effectively.

    Individuals could use AI to offer tailored advice on various personal finance topics. Additionally, AI has the potential to become an invaluable tool for financial advisors by enhancing their ability to analyze complex economic data, generate insights, and offer personalized recommendations to clients. The continued advancement of AI in this domain could lead to greater financial literacy and improved financial decision-making for both individuals and professionals.

    As LLMs become increasingly influential in personal finance, addressing key ethical concerns such as data privacy, algorithmic bias, and potential misuse is crucial. These models often require substantial personal information, underscoring the need for strong data protection measures and regulatory compliance. Algorithmic bias, often stemming from historical training data, can result in skewed financial advice, highlighting the importance of regular audits to ensure fairness. Additionally, preventing misuse—such as the spread of misleading or high-risk financial guidance—is vital to protect consumers.

    Rapid advancements in LLM capabilities reveal several promising areas for future research. Improving model interpretability is a key focus, as understanding the rationale behind AI-generated advice is essential for building user trust. Integrating real-time data is another priority. Because LLMs primarily rely on historical datasets, their advice can be outdated. Research incorporating live market trends and financial news can enhance the accuracy and timeliness of recommendations. Furthermore, exploring LLMs in behavioral finance can yield tailored insights, helping to address psychological biases like overconfidence or loss aversion. Enhancing these capabilities will enable more personalized, effective AI-driven financial advisory services.

    \section*{Contribution}
    \begin{itemize}
        \item Oudom Hean: conceptualization; project supervision; data analysis and interpretation; writing.
        \item Utsha Saha: conceptualization; data extraction; data cleaning; data analysis and interpretation; writing.
        \item Binita Saha: writing.
    \end{itemize}

    \section*{Funding}

    The project is not funded.

    \section*{Disclosure}

    The authors declare no conflicts of interest.

    \section*{Replication package}
    \begin{itemize}

        \item  Replication code is available upon request.
        \item MoneyCounts’ data were retrieved on July 18, 2024, from \url{https://financialliteracy.psu.edu/explore-a-financial-topic}.
        \item The National Financial Educators Council's (NFEC) data were retrieved on July 10, 2024 \url{https://www.financialeducatorscouncil.org/financial-literacy-test/}.
    \end{itemize}

    \printbibliography

\end{document}